# Title: Predicting Depressive Symptom Severity through Individuals' Nearby Bluetooth Devices Count Data Collected by Mobile Phones: A Preliminary Longitudinal Study

Yuezhou Zhang[1], MSc; Amos A Folarin[1,2,3], PhD; Shaoxiong Sun[1], PhD; Nicholas Cummins[1], PhD; Yatharth Ranjan[1], MSc; Zulqarnain Rashid[1], PhD; Pauline Conde[1], BSc; Callum Stewart[1], MSc; Petroula Laiou[1], PhD; Faith Matcham[4], PhD; Carolin Oetzmann[4], MSc; Femke Lamers[5], PhD; Sara Siddi[6,7,8], PhD; Sara Simblett[9], PhD; Aki Rintala[10,11], MSc; David C Mohr[12], PhD; Inez Myin-Germeys[10], PhD; Til Wykes[9], PhD; Josep Maria Haro[6,7,8], MD; Brenda WJH Pennix[5], PhD; Vaibhav A Narayan[13], PhD; Peter Annas[14], PhD; Matthew Hotopf[3,4], PhD; Richard JB Dobson[1,2,3], PhD; RADAR-CNS consortium[15]

[1]Department of Biostatistics & Health Informatics, Institute of Psychiatry, Psychology and Neuroscience, King's College London, London, United Kingdom
[2]Institute of Health Informatics, University College London, London, United Kingdom
[3]South London and Maudsley National Health Services Foundation Trust, London, United Kingdom
[4]Department of Psychological Medicine, Institute of Psychiatry, Psychology and Neuroscience, King's College London, London, United Kingdom
[5]Department of Psychiatry, Amsterdam Public Health Research Institute and Amsterdam Neuroscience, Amsterdam University Medical Centre, Vrije Universiteit and GGZ inGeest, Amsterdam, Netherlands
[6]Teaching Research and Innovation Unit, Parc Sanitari Sant Joan de Déu, Fundació Sant Joan de Déu, Barcelona, Spain
[7]Centro de Investigación Biomédica en Red de Salud Mental, Madrid, Spain
[8]Faculty of Medicine and Health Sciences, Universitat de Barcelona, Barcelona, Spain
[9]Department of Psychology, Institute of Psychiatry, Psychology and Neuroscience, King's College London, London, United Kingdom
[10]Center for Contextual Psychiatry, Department of Neurosciences, Katholieke Universiteit Leuven, Leuven, Belgium
[11]Faculty of Social Services and Health Care, LAB University of Applied Sciences, Lahti, Finland
[12]Center for Behavioral Intervention Technologies, Department of Preventive Medicine, Northwestern University, Evanston, Illinois, United States
[13]Janssen Research and Development LLC, Titusville, NJ, United States
[14] H. Lundbeck A/S, Copenhagen, Denmark
[15]The RADAR-CNS Consortium, London, United Kingdom

**Corresponding author**
Richard Dobson
Department of Biostatistics & Health Informatics
SGDP Centre, IoPPN
King's College London
Box PO 80
De Crespigny Park, Denmark Hill
London
SE5 8AF
Email: richard.j.dobson@kcl.ac.uk
Telephone: +44(0) 20 7848 0473

**Abstract**

**Background:** The Bluetooth sensor embedded in mobile phones provides an unobtrusive, continuous, and cost-efficient means to capture individuals' proximity information, such as the nearby Bluetooth devices count (NBDC). The continuous NBDC data can partially reflect individuals' behaviors and status, such as social connections and interactions, working status, mobility, and social isolation and loneliness, which were found to be significantly associated with depression by previous survey-based studies.

**Objective:** This paper aims to explore the NBDC data's value in predicting depressive symptom severity as measured via the 8-item Patient Health Questionnaire (PHQ-8).

**Method:** The data used in this paper included 2,886 bi-weekly PHQ-8 records collected from 316 participants recruited from three study sites in the Netherlands, Spain, and the UK as part of the EU RADAR-CNS study. From the NBDC data two weeks prior to each PHQ-8 score, we extracted 49 Bluetooth features, including statistical features and nonlinear features for measuring periodicity and regularity of individuals' life rhythms. Linear mixed-effect models were used to explore associations between Bluetooth features and the PHQ-8 score. We then applied hierarchical Bayesian linear regression models to predict the PHQ-8 score from the extracted Bluetooth features.

**Result:** A number of significant associations were found between Bluetooth features and depressive symptom severity. Generally speaking, along with the depressive symptoms worsening, one or more of the following changes were found in the preceding two weeks' NBDC data: (1) the amount decreased, (2) the variance decreased, (3) the periodicity (especially circadian rhythm) decreased, and (4) the NBDC sequence became more irregular. Compared with commonly used machine learning models, the proposed hierarchical Bayesian linear regression model achieved the best prediction metrics, $R^2 = 0.526$, and root mean squared error (RMSE) of 3.891. Bluetooth features can explain an extra 18.8% of the variance in the PHQ-8 score relative to the baseline model without Bluetooth features ($R^2=0.338$, RMSE = 4.547).

**Conclusion:** Our statistical results indicate that the NBDC data has the potential to reflect changes in individuals' behaviors and status concurrent with the changes in the depressive state. The prediction results demonstrate the NBDC data has a significant value in predicting depressive symptom severity. These findings may have utility for mental health monitoring practice in real-world settings.

## Introduction

The mobile phone has become a ubiquitous personal sensing device, with about 6 billion mobile phone users worldwide as of December 2020 [1]. It provides an unobtrusive, continuous, and cost-efficient means to capture individuals' daily behaviors using a number of embedded sensors, such as accelerometers, global positioning system (GPS) sensors, and Bluetooth sensors [2]. The embedded Bluetooth sensor can be used to record individuals' proximity information, such as the nearby Bluetooth devices count (NBDC) [3]. The continuously recorded NBDC data represents a mixed signal that could partially reflect individuals' behaviors and status, such as social connections and interactions [4-7], working status [8], mobility [9], and social isolation and loneliness [10].

Most of these individuals' behaviors and status were found to be significantly associated with depression by previous survey-based studies [11-16]. Individuals reporting lower social network connections or less social support tend to have higher depressive symptomatology [11]. Increased social support has positive effects on depression recovery, and poor social support may increase the risk of depression relapse [12]. The unemployment rate in depression is high, and depressive mood and medical comorbidity significantly influenced work status [13]. Reduced mobility and physical activity are associated with depressive symptoms [14]. Loneliness is a specific risk factor for depression, and a significant proportion of suicides have a history of social isolation [11,15].

Therefore, the NBDC data has the potential to reflect changes in people's behaviors and status during the depressive state. Compared with survey-based studies, the passively and continuously recorded NBDC data has several advantages, such as not relying on subjective recall and capturing dynamic information [6].

However, there have been studies exploring the relationship between the NBDC data and depression directly. Wang et al. found a negative association (r=-0.362, $P$=0.025) between the NBDC and self-reported depressive symptoms on the StudentLife dataset, which contains mobile phone data from 48 students across a 10-week term at Dartmouth College [16]. Boonstra et al. illustrated the feasibility of collecting nearby Bluetooth devices information for depression recognition tasks, but they did not provide further findings [6].

Several recent studies have investigated the relationships between Bluetooth proximity data and mental health [17-19]. Moturu et al. found that individuals with lower sociability (estimated by the NBDC) tend to report lower mood more often [17]. Bogomolov et al. established machine learning models to recognize happiness and stress with features of Bluetooth records, calls, and text messages, which obtained accuracy rates of 80.81% and 72.28%, respectively [18, 19]. The above three studies were all performed on the "Friends and Family" dataset, including 8 weeks of mobile phone data from 117 participants living in a major US university's married graduate student residency.

Previous studies [16-19] have been performed on relatively small (~100 participants), homogeneous (e.g., university students) cohorts of participants over relatively short periods (8-10 weeks), which may limit their generalizability. By contrast, our study was performed on a larger (N=316) multicenter dataset with longer follow-up (approximately 2 years). Bluetooth features used in previous studies [16-19] have been limited to basic statistical features (e.g., sum, mean and standard deviation), which are unable to characterize some nonlinear aspects (such as complexity and periodicity) of the Bluetooth data. These nonlinear characteristics can reflect individuals' life rhythms, such as circadian rhythms and social rhythms, which are affected by depressive symptoms [20]. Therefore, in this study, we leveraged multiscale entropy analysis [21] and frequency-domain analysis [22] to explore the nonlinear characteristics of the NBDC data.

In this paper, we aimed to explore the value of the NBDC data in predicting self-reported depressive symptom severity in a cohort of individuals with a history of recurrent major depressive disorder. Our first objective was to explore the associations between statistical Bluetooth features and depressive symptom severity. Our second objective was to extract nonlinear features quantifying complexity, regularity, and periodicity from the NBDC data and test their associations with depression. The third objective was to leverage appropriate machine learning models to predict the severity of depressive symptoms using extracted Bluetooth features.

## Method

### Dataset

The data used in this study were collected from a major EU Innovative Medicines Initiative (IMI) research programme Remote Assessment of Disease and Relapse - Central Nervous System (RADAR-CNS) [23]. The project aimed to investigate the use of remote measurement technologies (RMT) to monitor people with depression, epilepsy, and multiple sclerosis in real-world settings. The study protocol for the depression component (Remote Assessment of Disease and Relapse – Major Depressive Disorder; RADAR-MDD) has been described in detail by Matcham et al [24]. The RADAR-MDD project aimed to recruit 600 participants with a recent history of depression from three study sites in Spain (Centro de Investigación Biomédican en Red [CIBER]; Barcelona), the Netherlands (Vrije Universiteit Medisch Centrum [VUmc], Amsterdam]) and the UK (King's College London [KCL]). Recruitment procedures varied slightly across sites with eligible participants identified either through existing research infrastructures (in KCL and VUmc) where consent to be contacted for research purposes exists; through advertisements in general practices, psychologist practices and newspapers; through Hersenonderzoek.nl [25], a Dutch online registry (VUmc); through mental health services (in KCL and CIBER) [24].

Participants were asked to install passive and active remote monitoring technology (pRMT and aRMT, respectively) apps and use an activity tracker for up to 2 years of follow-up. Many categories of passive and active data were collected and uploaded to an open-source platform, RADAR-base [26].

As the purpose of this paper was to explore the value of the nearby Bluetooth devices count (NBDC) data in predicting self-reported depressive symptom severity, we focus on the NBDC data, Patient Health Questionnaire 8-item (PHQ-8) data [27], and baseline demographics. However, according to our previous research, the COVID-19 pandemic and related lockdown policies greatly impacted the behaviors (particularly mobility, social interactions, and working environment [working from home]) of European people [28]. To exclude the impact of the COVID-19 pandemic, we performed a preliminary analysis with the data before February 2020.

### PHQ-8 data

The variability of each participant's depressive symptom severity was measured via the PHQ-8, conducted by mobile phone every two weeks. The PHQ-8 score ranges from 0 to 24 (increasing severity) [27]. According to the PHQ-8 score, the severity of depression can usually be divided into 5 levels: asymptomatic (PHQ-8 < 5), mild (5≤ PHQ-8 <10), moderate (10≤ PHQ-8 <15), moderately severe (15≤ PHQ-8 <20), and severe (PHQ-8 ≥ 20) [27].

**Nearby Bluetooth devices count data**

The RADAR-base pRMT app scanned other Bluetooth devices in the participant's physical proximity once every hour. To avoid privacy leaks from participants and passers, the Media Access Control (MAC) address and types of Bluetooth devices were not recorded in this study. The nearby Bluetooth devices count (NBDC) was uploaded to the RADAR-base platform for further analyses.

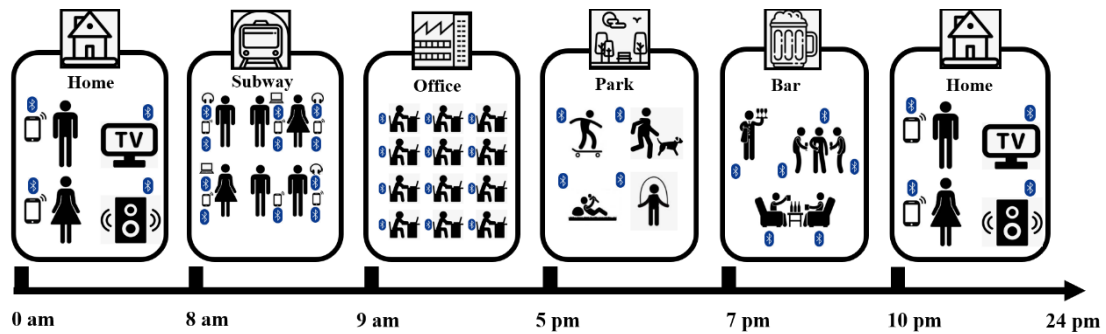

**Figure 1. A schematic diagram showing an individual's Nearby Bluetooth devices count (NBDC) in different scenarios in daily activities and life.**

Figure 1 is a schematic diagram showing an individual's NBDC in different scenarios in daily activities and life. At home, the NBDC is related to the number of family members and Bluetooth devices in the house, reflecting the participant's connections with family (whether living alone) and the number of other Bluetooth devices. In public transportation (such as train, subway, and bus), the NBDC is affected by the number of surrounding passengers' Bluetooth devices, reflecting participant's social connections with strangers. Studies have shown that whether feeling comfortable in the presence of strangers is related to the intensity of social connections [29]. In the company, the NBDC can reflect the participant's social connections and interactions with co-workers. After work, the NBDC can reflect whether the participant joins other social activities, such as going to the park or bar. Therefore, the NBDC data contains information about participants' social connections and interactions with family, friends, co-workers, and strangers, and it can also reflect participant's time at home, mobility, social isolation, working status, and the number of other Bluetooth devices in the house and working environment.

Figure 2 shows an example of two NBDC sequences collected over 14 days (336 hours) before two PHQ-8 records from one participant at two different depression severity levels (mild vs moderately severe).

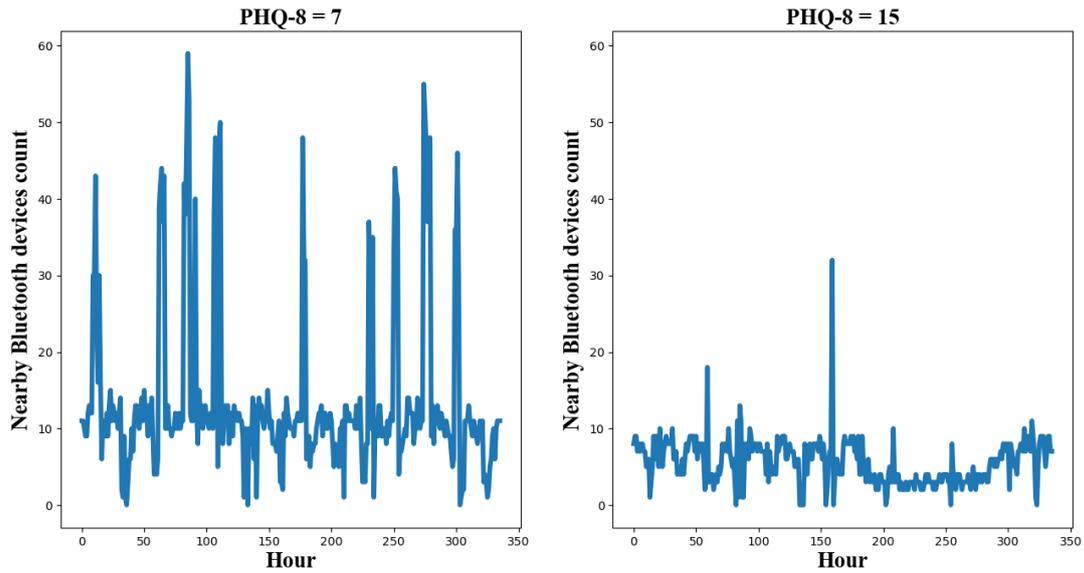

**Figure 2. An example of two 14-days nearby Bluetooth devices count (NBDC) sequences from the same participant at the mild depression level (left) and moderately severe level (right).**

## Demographics

Participants' demographics were recorded during the enrollment session. According to previous studies [30, 31], baseline age, gender, and education level were considered as covariates in our analyses. Due to the different educational systems in the three countries in our dataset, we used the number of years in education to represent education level.

## Data inclusion criteria and data preprocessing

For each PHQ-8 record, we considered a "PHQ-8 interval" that is 14 days before the day when the participant fills in the PHQ-8 questionnaire, as the PHQ-8 score used to represent the depressive symptom severity of the participant for the past two weeks. To reduce the impact of the COVID-19 pandemic and missing data on our analysis, we specified the following two data inclusion criteria:

(1) As mentioned in the dataset section, to exclude the impact of the COVID-19 pandemic, we restricted our analysis to PHQ-8 records prior to February 2020.

(2) Saeb et al. [32] and Farhan et al. [33] used 50% as each day's completeness threshold for passive data. In our dataset, 89.62% of days have 50% (12 hours) or more of the NBDC data. We considered one day as a "valid day" if it contains at least 12 hours' NBDC data. Then, we empirically selected PHQ-8 intervals with at least 10 valid days as valid PHQ-8 intervals to retain the majority (81.78%) of PHQ-8 intervals.

For the NBDC sequence in each selected PHQ-8 interval, we used linear interpolation to impute the missing hours in all valid days and discarded the NBDC data that did not belong to a valid day. The "NBDC sequence" in the rest of this paper refers to the pre-processed NBDC data in the 14-days PHQ-8 interval.

## Feature extraction

According to past Bluetooth-related research [16-19] and other research from the field of signal processing [21, 22], we extracted 49 Bluetooth features from the NBDC sequence in the PHQ-8 interval in the following three categories: second-order statistics, multiscale entropy, and frequency-domain. Table 1 summarizes all Bluetooth features extracted in this paper.

## Second-order statistical features

We first calculated 4 daily features (max, min, mean and standard deviation) of daily NBDC data from all valid days in the PHQ-8 interval. For each daily feature, we calculated 4 second-order features (max, min, mean and standard deviation) to reflect the amount and variance of the NBDC in the PHQ-8 interval. These features were denoted in the following format: [Second-order feature] _ [Daily feature]. For example, the average value of the daily maximum number of the NBDC in the PHQ-8 interval was denoted as *Mean_Max*. A total of 16 second-order statistical features were extracted.

## Nonlinear Bluetooth features

The second-order statistical features can only reflect the amount (Max, Min, Mean) and variance (Standard deviation) of the NBDC data. To exploit more information embedded in the NBDC data, we proposed multiscale entropy (MSE) and frequency domain (FD) features to measure the nonlinear characteristics, such as regularity, complexity and periodicity, of the NBDC sequence.

### Multiscale entropy features

Multiscale entropy (MSE) analysis has been used to provide insights into the complexity and periodicity of signals over a range of timescales since the method was proposed by Costa et at. [21]. It has been widely used in the field of signal analysis, such as heart rate variability analysis [34], electroencephalogram analysis [35] and gait dynamics analysis [36]. Compared with other entropy techniques (e.g., sample entropy and approximate entropy), the advantage of MSE analysis is the assessments of complexity at shorter and longer timescales can be analyzed separately [37]. The MSE at short timescales reflects the complexity of the sequence. The larger the MSE at short timescales, the more chaotic and irregular the signal. The MSE at relatively long timescales assesses fluctuations occurring at a certain period, reflecting the periodicity of the signal.

To explore the complexity and periodicity of the NBDC sequence on different timescales (from 1 hour to 24 hours), we calculated MSE features of the NBDC sequences from scale 1 to scale 24, denoted as *MSE_1, MSE_2, …, MSE_24*, respectively. Figure 3 shows an example of MSE features calculated on two NBDC sequences at different depression severity levels from the same participant shown in Figure 2. In this example, the NBDC sequence at the mild depression level (PHQ-8=7) has lower MSE at relatively short timescales (scale 1-3) and higher MSE at relatively long timescales than the sequence at the moderately severe depression level (PHQ-8=15). This indicated that this participant's NBDC sequence at the mild depression level was more regular and periodic than the NBDC sequence at the moderately severe depression level.

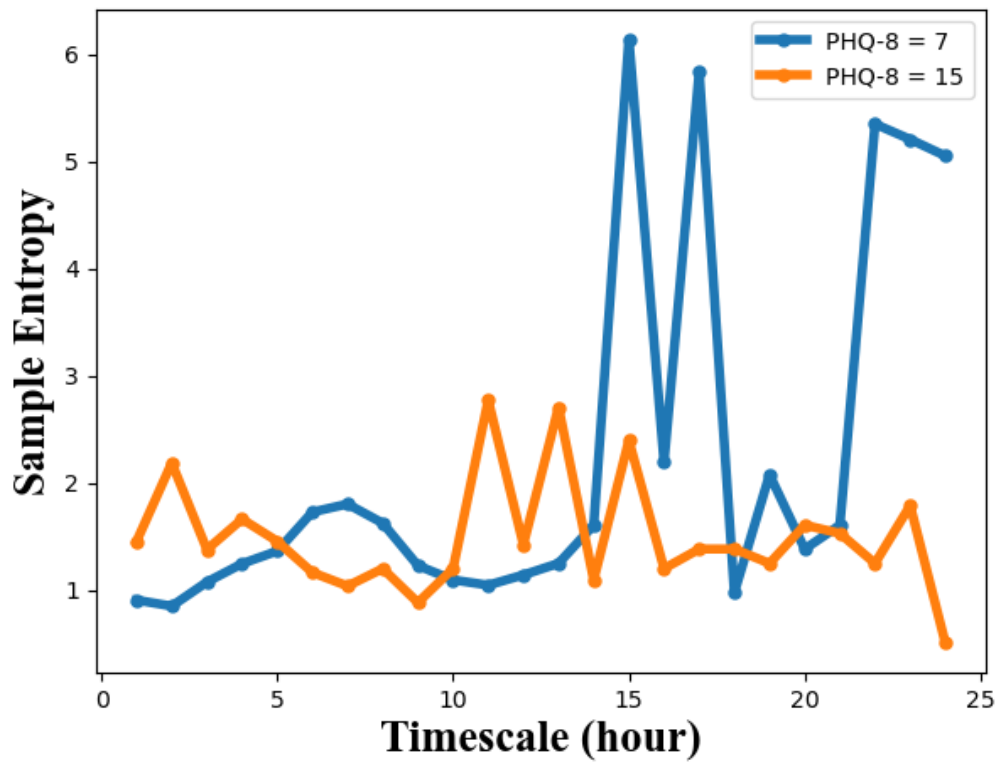

**Figure 3.** An example of multiscale entropy (scale 1- 24) of two 14-days nearby Bluetooth device count (NBDC) sequences at the mild depression level (blue) and the moderately severe level (orange) from the same participant in Figure 2.

### Frequency domain features

Frequency-domain (FD) analysis has been widely used in the signal processing field, especially for signals with periodic characteristics [22]. People's behaviors follow a quasi-periodic routine such as sleeping at night, working on weekdays, and gathering with friends on weekends [20, 38]. We therefore leveraged FD analysis to explore the periodic patterns in the NBDC data. The fast Fourier transformation (FFT) was performed to transform the NBDC sequence from the time domain to the frequency domain. We set the sample rate to 24 hours, then the spectrum generated by FFT had the frequency axis scaled to reflect cycles/day.

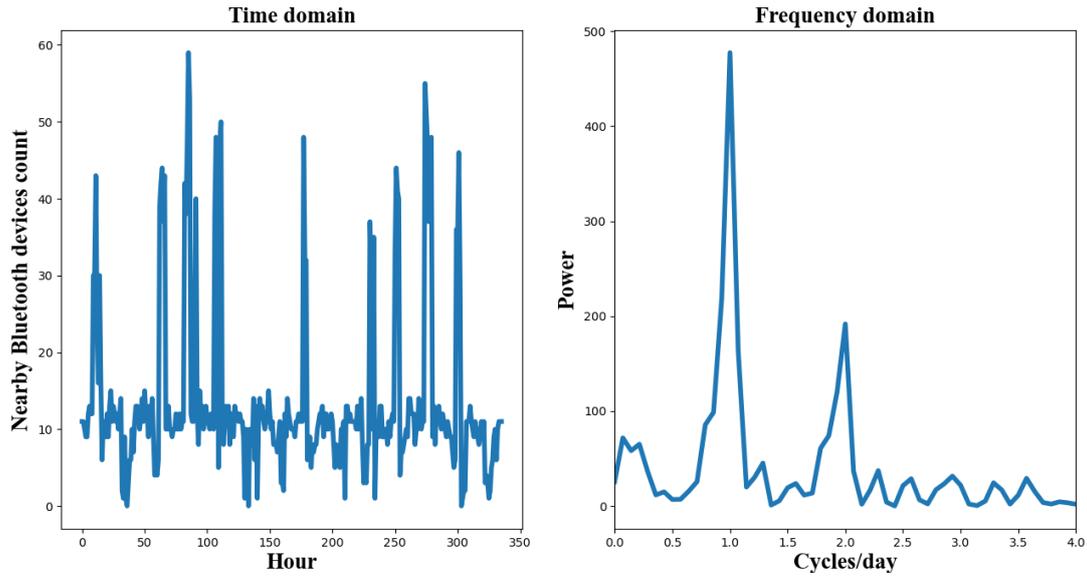

**Figure 4. An example of a 14-days NBDC sequence in the time domain (left) and its spectrum in the frequency domain (right).**

Figure 4 is an example of a NBDC sequence in the time domain and its spectrum in the frequency domain. According to the spectrum's definition, spectrum power around 1 cycle/day reflects the participant's circadian rhythm (~24-hour rhythm) [20]. To explore the periodic rhythms of different period lengths, we empirically defined 3 frequency intervals: low frequency (LF) (0-0.75 cycles/day), middle frequency (MF) (0.75-1.25 cycles/day) and high frequency (HF) (>1.25 cycles/day). The power in MF represents the circadian rhythm. Similarly, the power in LF represents the long-term (> 1 day) rhythm, while HF's power indicates the short-term (< 1 day) rhythm.

The sums of spectrum power in these 3 frequency intervals were calculated and denoted as *LF_sum, MF_sum, HF_sum*, respectively. The percentages of spectrum power in these 3 frequency intervals to the total spectrum power were extracted and denoted as *LF_pct, MF_pct, and HF_pct*. To estimate the complexity and regularity of the spectrum, we calculated spectral entropy (SE) [39] in these 3 intervals, denoted as *LF_se, MF_se*, and *HF_se*, respectively.

**Table 1. A summary of 49 Bluetooth features used in this paper and their short descriptions.**

| Category | Abbreviation | Description | Number of features |
| --- | --- | --- | --- |
| Statistical features | [Second-order feature]_[Daily feature] e.g., Max_Mean | Second-order features (max, min, mean, and standard deviation) calculated in the PHQ-8 interval based on daily statistical Bluetooth features (max, min, mean, and standard deviation). | 16 |
| Multiscale entropy | MSE_1, MSE_2, …, MSE_24 | Multiscale entropy of the NBDC[a] sequences from scale 1 to scale 24. | 24 |

| Frequency domain | LF_sum, MF_sum, HF_sum | The sums of spectrum power in LF, MF, and HF. | 3 |
|---|---|---|---|
| Frequency domain | LF_pct, MF_pct, HF_pct | The percentages of spectrum power in LF, MF, and HF[b] to the total spectrum power | 3 |
| Frequency domain | LF_se, MF_se, HF_se | Spectral entropy in LF, MF, and HF. | 3 |
| Total | | | 49 |

[a] NBDC: nearby Bluetooth devices count.
[b] LF: low frequency (0-0.75 cycles/day); MF: middle frequency (0.75-1.25 cycles/day); HF: high frequency (>1.25 cycles/day).

**Statistical methods**

The linear mixed-effect model contains both fixed and random effects, allowing for both within-participant and between-participants variations over repeated measurements [40]. Therefore, we used linear mixed-effect models in our statistical analyses.

### Pairwise association analyses

To explore the association between each Bluetooth feature and depression severity, a series of pairwise linear mixed-effect models with random participant intercepts were performed to regress the PHQ-8 score with each of Bluetooth features. All mixed-effect models, baseline age, gender, and years in education were considered as covariates. The z-test was used to evaluate the statistical significance of the coefficient of each model. The Benjamini-Hochberg method [41] was used for multiple comparisons correction, and the significant level for adjusted *P*-value was set to 0.05. All linear mixed-effect models were implemented by using the R package ("lmerTest"), and the Benjamini-Hochberg method was performed by using the command "p.adjust" in R software.

### Likelihood ratio test

One objective of this paper was to assess what value these Bluetooth features provide beyond other information that might be readily available, such as baseline demographics. The likelihood ratio test is a statistical test of goodness-of-fit between two nested models [42]. If the model with more parameters fit the data significantly better, it indicated additional parameters provided more information and improved the model's fitness [42]. Therefore, we built three nested linear mixed-effect models with random participant intercepts (Model A, Model B, and Model C). The predictors of Model A were only demographics. The predictors of Model B were demographics and 16 second-order statistical features. The predictors of Model C were demographics and all 49 Bluetooth features. The likelihood ratio tests were performed to test whether these Bluetooth features have a significant value in fitting the PHQ-8 score regression model.

### Prediction models

Another objective of this paper was to examine whether it is possible to predict participants' depressive symptom severity using Bluetooth features combined with some known information (demographics and previous PHQ-8 scores).

## Hierarchical Bayesian linear regression model

The hierarchical Bayesian approach is an intermediate method compared to the completely pooled model and individualized model, capturing the whole population's characteristics while allowing individual differences [43]. We leveraged the hierarchical Bayesian linear regression model to predict participants' PHQ-8 score using Bluetooth features, demographics (age, gender, and years in education), and the last observed PHQ-8 score. In this study, we implemented the hierarchical Bayesian linear regression using the "PyMC3" package [44] in Python. To compare the results with other commonly used machine learning models, we also implemented the LASSO regression model [45] and XGBoost regression model [46] using the Scikit-learn machine learning library [47] in Python. As depressive mood has a strong autocorrelation [48], we considered a baseline hierarchical Bayesian linear regression model with the last observed PHQ-8 score and demographics as predictors.

## Model evaluation

We selected root mean squared error (RMSE) and the predicted coefficient of determination ($R^2$) as two metrics for model discrimination evaluation. As we used the temporal data, "future data" should not predict "past data". Therefore, only the data observed before test data can be included in the training set. We applied leave-all-out (LAO) and leave-one-out (LOO) time-series cross-validation described in [48]. As the number of PHQ-8 intervals of each participant in our data is different, we made some minor modifications to these two schemes (Figure 5).

(1) LAO time-series cross-validation: Each participant's data were divided into a sequence of t consecutive same-sized test sets, where the size of each test set is the length of one PHQ-8 interval (14 days), and t is the number of PHQ-8 intervals of this participant. The corresponding training set included all PHQ-8 intervals before each test set. Then test sets and training sets were pooled across all participants. This process generated T-1 test and training set pairs (no prior data to predict the first PHQ-8 score), where T is the maximum number of PHQ-8 intervals of one participant in our dataset ($t \leq T$).

(2) LOO time-series cross-validation: Each participant's data were divided into a training set and a test set. The training set was constructed using the first two PHQ-8 intervals of a participant, with the test set containing the rest of the participant's PHQ-8 intervals. Then the training set was pooled with all data from all other participants. This scheme generated J training and test set pairs, where J is the number of participants in our dataset.

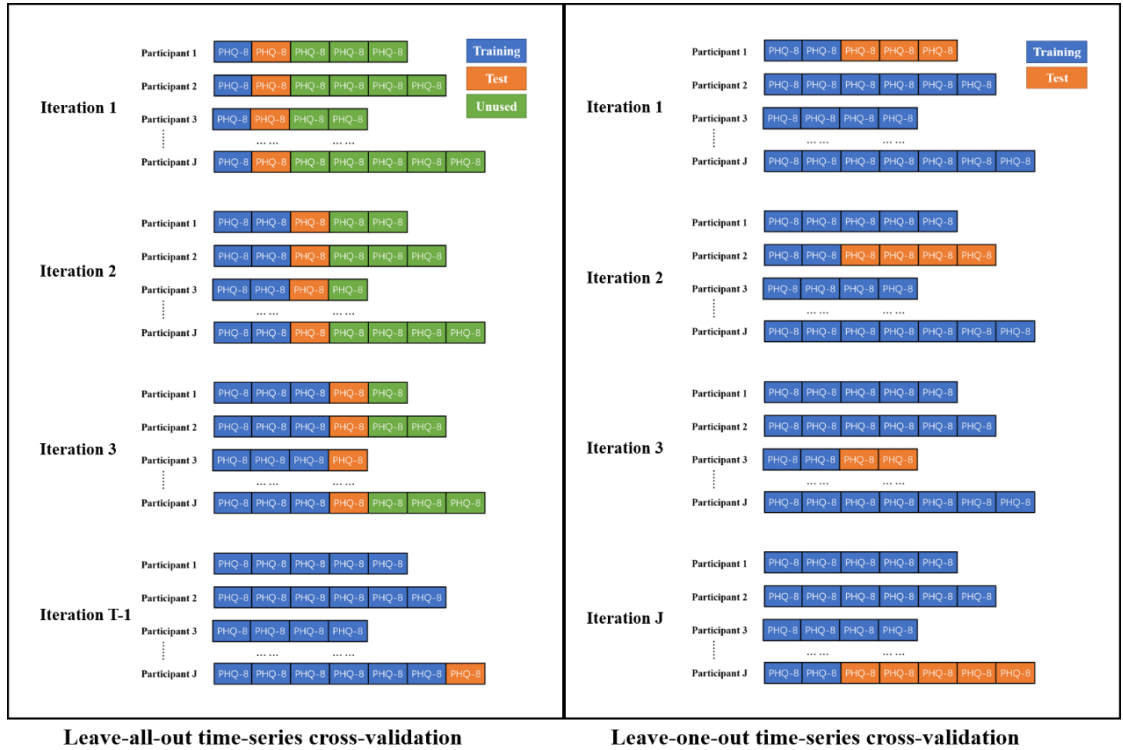

**Figure 5.** Two schematic diagrams of leave-all-out time-series cross-validation (left) and leave-one-out time-series cross-validation (right), where T is the maximum number of PHQ-8 intervals of one participant, J is the number of participants, the training set is indicated by blue, the test set is indicated by orange, and unused data is indicated by green.

## Results

### Data summary

According to our date inclusion criteria, from June 2018 to February 2020, 2886 PHQ-8 intervals from 316 participants collected from three study sites were selected for our analysis. Table 2 presents a summary of demographics and distribution of PHQ-8 records of all selected participants.

**Table 2. A summary of demographics and the PHQ-8[a] records distribution of all selected participants.**

| Characteristic | Statistics |
|---|---|
| Number of participants | 316 |
| **Demographics** | |
| Age at baseline, median (Q1, Q3) | 51.0 (35.0, 59.0) |
| Female sex, n (%) | 234 (74.05%) |
| Number of years in education, median (Q1, Q3) | 16.0 (14.0, 19.0) |
| **PHQ-8 records distribution** | |
| Number of PHQ-8 intervals | 2886 |

| | | |
|---|---|---|
| | Number of PHQ-8 intervals for each participant, median (Q1, Q3) | 8.0 (3.0, 14.0) |
| | PHQ-8 scores, median (Q1, Q3) | 9.0 (5.0, 15.0) |

[a] Patient Health Questionnaire 8-item

**Association analysis results**

The significant associations between depression severity (the PHQ-8 score) and Bluetooth features are presented in Table 3. Figure 6 represents pairwise Spearman correlation coefficients between all 49 Bluetooth features.

**Associations between the PHQ-8 score and second-order statistical features**

There were 10 second-order statistical features significantly associated with the PHQ-8 score. All these significant associations were negative, i.e., the larger the value of these features, the lower the PHQ-8 score. Notably, *Min_Max* (the minimum value of daily maximum NBDC in past 14 days) had the strongest association ($z = -4.431$, $P < .001$), which indicated that participants with lower PHQ-8 score tended to have more daily social activities (such as social interactions and traveling) in the past two weeks. In addition, 4 features related to daily variance (*Max_Std, Min_Std, Mean_Std,* and *Std_Std*) of the NBDC were all significantly and negatively associated with depression. The possible reason is that participants with lower PHQ-8 scores were more inclined to visit different places (high mobility) each day, so their daily variance of the NBDC was higher.

**Associations between the PHQ-8 score and multiscale entropy features**

Multiscale entropy (MSE) at scale 1, scale 2, and scale 3 (*MSE_1, MSE_2,* and *MSE_3*) were significantly and positively associated with the PHQ-8 score, while MSE at scale 16 and scale 22 (*MSE_16* and *MSE_22*) were significantly and negatively associated with depressive symptom severity. According to the explanations of MSE we mentioned in the method section, these associations indicated that the participants with more irregular and chaotic NBDC sequences were likely to have more severe depressive symptoms, while those with periodic and regular NBDC sequences may have lower PHQ-8 scores.

**Associations between the PHQ-8 score and frequency-domain features**

There were 5 frequency-domain (FD) features significantly associated with the PHQ-8 score. The spectrum power is related to both the amount and frequency component of the NBDC sequence, so it had relatively strong correlations with second-order statistical features (Figure 6). Therefore, the spectrum power of three frequency intervals (*LF_sum, MF_sum,* and *HF_sum*) were all significantly and negatively associated with the PHQ-8 score. Among them, the *MF_sum* had the strongest association ($z = -4.766$, $P < .001$) with depression, which indicated the circadian rhythm of the NBDC sequence is important to reflect the severity of depression. Likewise, the percentage of middle-frequency power (*MF_pct*) was significantly and negatively associated with depressive symptom severity. The spectral entropy of high-frequency (*HF_se*) was significantly and positively associated with depression. This indicated that the participants with irregular short-term (< 1 day) rhythms were likely to have more severe depressive symptoms.

**Table 3. Coefficient estimates, standard error (SE), z-test statistics, and *P* values from pairwise linear mixed-effect models for exploring associations between Bluetooth features and the depressive symptom severity (PHQ-8[a] score). (Only significant associations [adjusted *P*-value < 0.05] are reported.)**

| Feature | | Estimates | SE | z score | Adjusted *P*-value[b] |
|---|---|---|---|---|---|
| **Second-order statistical** | | | | | |
| | Min_Max | -0.052 | 0.012 | -4.431 | <.001 |
| | Mean_max | -0.016 | 0.006 | -2.809 | .005 |
| | Max_std | -0.015 | 0.006 | -2.657 | .008 |
| | Min_Std | -0.215 | 0.056 | -3.838 | <.001 |
| | Mean_Std | -0.065 | 0.023 | -2.802 | .005 |
| | Std_Std | -0.048 | 0.020 | -2.385 | .02 |
| | Max_Mean | -0.030 | 0.008 | -3.498 | <.001 |
| | Min_Mean | -0.093 | 0.046 | -2.036 | .04 |
| | Mean_Mean | -0.083 | 0.026 | -3.225 | .001 |
| | Std_Mean | -0.095 | 0.027 | -3.464 | .001 |
| **Multiscale entropy** | | | | | |
| | MSE_1 | 0.642 | 0.225 | 2.853 | .005 |
| | MSE_2 | 0.433 | 0.192 | 2.255 | .02 |
| | MSE_3 | 0.401 | 0.202 | 1.985 | .04 |
| | MSE_16 | -0.102 | 0.042 | -2.429 | .01 |
| | MSE_22 | -0.123 | 0.043 | -2.860 | .005 |
| **Frequency-domain** | | | | | |
| | LF_sum | -2.060 | 0.533 | -3.865 | <.001 |
| | MF_sum | -6.720 | 1.410 | -4.766 | <.001 |
| | HF_sum | -2.710 | 1.040 | -2.606 | .009 |
| | MF_pct | -1.834 | 0.812 | -2.259 | .02 |
| | HF_se | 3.821 | 1.820 | 2.099 | .04 |

[a] Patient Health Questionnaire 8-item
[b] *P*-values were adjusted by the Benjamini-Hochberg method for multiple comparisons correction.

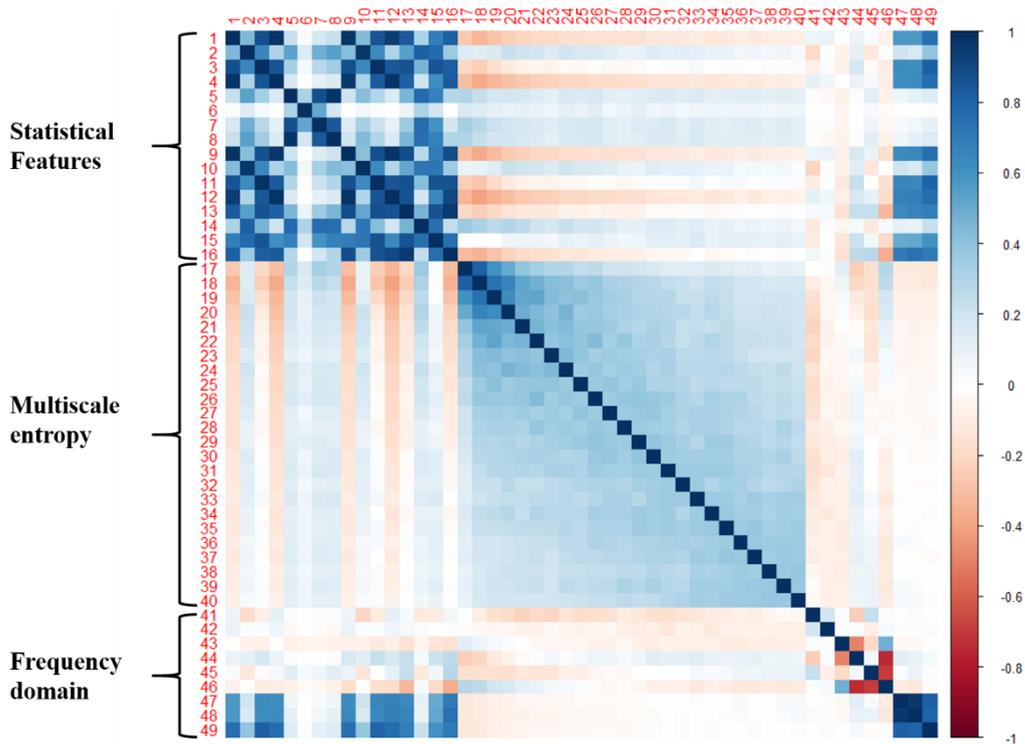

**Figure 6. A correlation plot of pairwise Spearman correlations between all 49 Bluetooth features.**

### The results of likelihood ratio tests

The results of the likelihood ratio tests are presented in Table 4. Model B (with second-order statistical Bluetooth features) and Model C (with all Bluetooth features) both fitted data significantly better than Model A (without Bluetooth features), which indicated Bluetooth features could improve the statistical model significantly. The goodness-of-fit of Model C was significantly better than Model B, indicating nonlinear Bluetooth features (MSE and FD features) provided additional information to the statistical model.

**Table 4. The results likelihood ratio tests of three nested linear mixed-effect models.**

| Model | Diff. of parameters[d] | $\chi^2$ [e] | P-value |
|---|---|---|---|
| Model B[b] vs Model A[a] | 16 | 31.04 | .01 |
| Model C[c] vs Model A | 49 | 135.19 | <.001 |
| Model C vs Model B | 33 | 104.15 | <.001 |

[a] Predictors of Model A: demographics.
[b] Predictors of Model B: demographics + 16 second-order statistical features.
[c] Predictors of Model C: demographics + 16 second-order statistical features + 24 multiscale entropy features + 9 frequency-domain features.
[d] Difference of parameters
[e] The critical values of the likelihood ratio statistic are: $\chi^2_{0.05}(16) = 26.296$, $\chi^2_{0.05}(33) = 47.400$, and $\chi^2_{0.05}(49) = 66.339$.

**Performance of prediction models**

For the prediction models, a subset of 183 participants was selected based on the following rules:

(1) To ensure that each participant had sufficient PHQ-8 intervals for the LOO time-series cross-validation (described in the model evaluation part), the number of valid PHQ-8 intervals for each participant should be at least 3.
(2) To test whether the model can predict variability of depression severity, the difference of one participant's PHQ-8 scores should be more than or equal to 5 (clinically meaningful change) [49].

The results of the leave-all-out (LAO) and the leave-one-out (LOO) time-series cross-validation are presented in Table 5. The $R^2$ score of the baseline model was 0.338 in LAO time-series cross-validation, which showed that more than 30% variance could be explained by the last observed PHQ-8 score and baseline demographics. In LOO time-series cross-validation, the $R^2$ score of the baseline model was negative, which indicated the baseline model did not explain any variance in the LOO time-series cross-validation. To assess the improvement from nonlinear Bluetooth features, we tested the hierarchical Bayesian model with and without nonlinear Bluetooth features separately.

In the subset, the maximum number of PHQ-8 intervals of one participant was 27, so the LAO time-series cross-validation went through T-1=26 iterations. The hierarchical Bayesian linear regression model with all Bluetooth features achieved the best result ($R^2$ =0.526, RMSE = 3.891), beating the LASSO and XGBoost regression models. Compared with the result of the baseline model ($R^2 = 0.338$), the improvement in $R^2$ score is 0.188, which means the Bluetooth features explained an additional 18.8% of data variance. The nonlinear Bluetooth features explained an additional 4.5% of data variance in the hierarchical Bayesian model.

The number of subset participants was 183, so J=183 iterations of the LOO time-series cross-validation were performed. The hierarchical Bayesian linear model with all Bluetooth features had the best performance ($R^2 = 0.387$, RMSE =4.426), but is close to the result of the XGBoost regression model ($R^2 = 0.346$ RMSE = 4.523).

The performance of the hierarchical Bayesian linear regression model evaluated by the LAO cross-validation was better than the LOO cross-validation performance. One potential reason is that only the first two PHQ-8 intervals of one participant were used for training in the LOO cross-validation, which may cause the model to underfit the patterns at the participant level.

**Table 5. Results of the leave-all-out time-series cross-validation (left) and leave-one-out time-series cross-validation (right) of the hierarchical Bayesian linear regression model, commonly used machine learning models and the baseline model.**

| Model | Leave-all-out | | Leave-one-out | |
|---|---|---|---|---|
| | $R^2$ | RMSE | $R^2$ | RMSE |
| Baseline model[a] | 0.338 | 4.547 | -0.074 | 5.802 |
| LASSO regression | 0.458 | 4.114 | 0.144 | 5.178 |
| XGBoost regression | 0.464 | 4.092 | 0.346 | 4.523 |
| Hierarchical Bayesian linear (second-order statistical features) | 0.481 | 4.026 | 0.353 | 4.501 |
| Hierarchical Bayesian linear (all Bluetooth features) | 0.526 | 3.891 | 0.387 | 4.426 |

[a] The baseline model is the hierarchical Bayesian linear regression model with only the last observed PHQ-8 score and demographics as predictors.

## Discussion

### Principal findings

To explore the value of the nearby Bluetooth devices count (NBDC) data in predicting depression severity, we extracted 49 features from the NBDC sequences in the following three categories: second-order statistical features, multiscale entropy (MSE) features, and frequency-domain (FD) features. To the best of the authors' knowledge, this is the first time that MSE and FD features have been used in the NBDC and depression data analysis. According to the results of association analyses (Table 3), when depression symptoms worsen (increasing in the PHQ-8 score), one or more following changes were seen in the preceding 14 days' NBDC sequence: (1) the amount decreased, which is consistent with Wang et al.'s finding in [16], (2) the variance decreased, (3) the periodicity (especially circadian rhythm) decreased, and (4) the NBDC sequence became more irregular and chaotic.

These changes in the NBDC data can be explained by depression symptoms. The main manifestations of depression include negative feelings (such as sadness, guilt, stress, and tiredness) and loss of interest or pleasure [50]. This may lead to the following changes in behaviors, such as the increased time at home [32, 51,52], decreased mobility [14, 32], loss of the ability to work or study [13, 50], reduced intensity of social interactions [11], unstable and irregular sleep [53], and decreased engagement in activities [54]. The increased time at home, inability to work or study, and diminished social interactions are reflected in the reduced amount of the NBDC sequence. The decreased mobility and engagement in activities may reduce the variance of the NBDC sequence. Depression also may lead to misalignment of circadian rhythms and make people's life rhythms (such as sleep rhythms and social rhythms) more irregular [20]. This can be reflected in reduced periodicity and increased irregularity of the NBDC sequence. Saeb et al. [32] and Farhan et al. [33] found similar findings in GPS data which showed the circadian rhythms of GPS signal was significantly and negatively correlated with depression.

From the perspective of the statistical model, Bluetooth features extracted in this paper significantly improved the goodness-of-fit for the PHQ-8 score, and nonlinear Bluetooth features (MSE and FD features) can provide additional information to second-order statistical features (Table 4). From the perspective of the prediction model, these 49 Bluetooth features explained an extra 18.8% of the variance in the PHQ-8 score relative to the baseline model, containing only the last PHQ-8 score and demographics, and MSE and FD features explained an extra 4.5% of data variance in the hierarchical Bayesian model (Table 5). From the perspective of the correlations between Bluetooth features (Figure 6), we can observe that, except for 3 FD features related to the spectrum power that had relatively strong correlations with second-order statistical features, the correlations between other nonlinear Bluetooth features and second-order statistical features were not obvious. This indicated that the MSE and FD features capture dimensions of information to second-order statistical features.

In our prediction model, the hierarchical Bayesian linear regression model achieved the best results in both the leave-all-out (LAO) and leave-one-out (LOO) time-series cross-validation. Compared with other models, one of the advantages of the hierarchical Bayesian model is that it performs individual predictions while considering the population's common characteristics [43]. Therefore, the hierarchical Bayesian model can be considered a suitable prediction modelling method for longitudinal data. The LOO time-series cross-validation results illustrated that the hierarchical Bayesian model could predict depression for participants with few observations (only 2 PHQ-intervals in the training set) that overcomes the cold start problem. The hierarchical Bayesian linear model achieved a better result in the LAO time-series cross-validation, which indicated that the prediction results gradually become more accurate and individualized when each participant has more data available in the training set.

**Limitations**

The RADAR-MDD project was designed for long-term monitoring (up to 2 years) and collecting many other passive data, such as GPS, acceleration, app usage, and screen lightness, which need to be collected simultaneously through the mobile phone. Therefore, to avoid excessive battery consumption, the nearby Bluetooth devices were scanned hourly in this study. However, some past studies suggested scanning the nearby Bluetooth devices every 5 minutes to achieve high enough temporal resolution [5, 19]. Although hourly NBDC data can also reflect individuals' behaviors and status, our lower data resolution may cause the loss of some dynamic information. On the other hand, using the relatively low resolution enables us to collect multimodal data without excessive battery consumption. As we mentioned that the NBDC data is related to individuals' movement and location information, we will combine the NBDC data with GPS and acceleration data for future analysis to understand the context of the Bluetooth data.

As we mentioned in the method section, the MAC address and types of Bluetooth devices were not recorded for private issues. This made it impossible to distinguish between mobile phones and other Bluetooth devices (such as headphones, printers, and laptops), nor can it distinguish between strangers' and acquaintances' devices. The advantage of the NBDC data is that it contains mixed and rich information. The disadvantage is that it is difficult to explain the specific reasons for changes in the NBDC, i.e., we cannot know whether the changes in the NBDC are caused by social interactions, working status, traveling, or isolation. Therefore, this paper did not explain in depth the actual meaning behind the Bluetooth features. For this limitation, we plan to use the hashed MAC address in future research.

For the FD features, the division of the frequency intervals of the spectrum of the NBDC sequence in this paper was manually specified by experience. The purpose of extracting these FD features was to prove that the NBDC sequence's frequency domain has the potential to provide more information about individuals' behaviors and life rhythms. It is necessary to discuss the optimal boundaries of frequency intervals of the NBDC data in future research.

This paper applied the hierarchical Bayesian linear regression model to explore the linear relationships between Bluetooth features and depression. However, there may be nonlinear relationships between social connections and depressive symptom severity. Gaussian process [55], using the kernel method to find the nonlinear relationships, will be considered in future research.

# Conclusion

Our statistical results indicated that the nearby Bluetooth device counts (NBDC) data has the potential to reflect changes in individuals' behaviors and status during the depressive state. The prediction results demonstrated the NBDC data has a significant value in predicting depressive symptom severity. The nonlinear Bluetooth features proposed in this paper provided additional information to statistical and prediction models. The hierarchical Bayesian model is an appropriate prediction model for predicting depression with longitudinal data, as participant-level and population-level characteristics are both considered in the model. These findings may support mental health monitoring practice in real-world settings.

# Acknowledgments

The RADAR-CNS project has received funding from the Innovative Medicines Initiative 2 Joint Undertaking under grant agreement No 115902. This Joint Undertaking receives support from the European Union's Horizon 2020 research and innovation programme and EFPIA, www.imi.europa.eu. This communication reflects the views of the RADAR-CNS consortium and neither IMI nor the European Union and EFPIA are liable for any use that may be made of the information contained herein. The funding body have not been involved in the design of the study, the collection or analysis of data, or the interpretation of data.

Participants in the CIBER site came from following four clinical communities in Spain: Parc Sanitari Sant Joan de Déu Network services, Institut Català de la Salut, Institut Pere Mata, and Hospital Clínico San Carlos.

Participant recruitment in Amsterdam was partially accomplished through Hersenonderzoek.nl, a Dutch online registry that facilitates participant recruitment for neuroscience studies [25]. Hersenonderzoek.nl is funded by ZonMw-Memorabel (project no 73305095003), a project in the context of the Dutch Deltaplan Dementie, Gieskes-Strijbis Foundation, the Alzheimer's Society in the Netherlands and Brain Foundation Netherlands.

This paper represents independent research part funded by the National Institute for Health Research (NIHR) Maudsley Biomedical Research Centre at South London and Maudsley NHS Foundation Trust and King's College London. The views expressed are those of the author(s) and not necessarily those of the NHS, the NIHR or the Department of Health and Social Care.

We thank all the members of the RADAR-CNS patient advisory board for their contribution to the device selection procedures, and their invaluable advice throughout the study protocol design.

This research was reviewed by a team with experience of mental health problems and their careers who have been specially trained to advise on research proposals and documentation through the Feasibility and Acceptability Support Team for Researchers (FAST-R): a free, confidential service in England provided by the National Institute for Health Research Maudsley Biomedical Research Centre via King's College London and South London and Maudsley NHS Foundation Trust.

RADAR-MDD will be conducted per the Declaration of Helsinki and Good Clinical Practice, adhering to principles outlined in the NHS Research Governance Framework for Health and Social Care (2nd edition). Ethical approval has been obtained in London from the Camberwell St Giles Research Ethics Committee (REC reference: 17/LO/1154), in London from the CEIC Fundacio Sant Joan de Deu (CI: PIC-128-17) and in The Netherlands from the Medische Ethische Toetsingscommissie VUms (METc VUmc registratienummer: 2018.012 – NL63557.029.17)

## Conflicts of Interest